\documentclass[letterpaper, 10 pt, conference]{ieeeconf}
\IEEEoverridecommandlockouts    

\usepackage{multirow}
\usepackage{lipsum}
\usepackage{amsmath}
\usepackage{amssymb}
\usepackage[ruled,vlined]{algorithm2e}
\usepackage{graphicx}
\usepackage{booktabs}
\usepackage{cases}
\usepackage{geometry}
\usepackage{afterpage}
\graphicspath{{./Figures/}}
\usepackage[hidelinks]{hyperref}
\usepackage[caption=false]{subfig}
\usepackage{xcolor}
\usepackage{empheq}

\newcommand{\GG}{$g$-$g$}
\newcommand{\GGV}{$g$-$g$-$v$}

\newif\ifanonymous
\anonymoustrue

\title{\LARGE \bf Trajectory Planning and Control near the Limits:\\an Open Experimental Benchmark on the RoboRacer Platform} 

\author{
Mattia Piccinini$^{1}$, Patrick Zambiasi$^{2}$, Aniello Mungiello$^{3}$, Mattia Piazza$^{4}$, Felix Jahncke$^{1}$, Johannnes Betz$^{1}$
\thanks{$^{1}$ Professorship of Autonomous Vehicle Systems, Technical University
of Munich, 85748 Garching, Germany; Munich Institute of Robotics and Machine Intelligence (MIRMI).}%
\thanks{$^{2}$ Avilus GmbH, Germany.}%
\thanks{$^{3}$ Department of Information Technology and Electrical Engineering (DIETI), University of Naples Federico II, Naples 80125, Italy.}%
\thanks{$^{4}$ Dept. of Industrial Engineering,
University of Trento, 38123 Trento, Italy.}%
}

\begin{document}
	
	\maketitle
	
    \begin{abstract}
    We present a modular framework to benchmark new and existing methods for trajectory planning and control in high-acceleration maneuvers that push autonomous driving to the limits. Our framework includes time-optimal raceline generation, online time-optimal velocity replanning, geometric path tracking controllers, and a new model-structured neural network (MS-NN) to learn the inverse dynamics for steering control. We deploy our framework on a 1:10-scale RoboRacer platform, using two circuits. Through several ablations with cautious and aggressive racelines, we study the performance of single modules and their combinations. We show that our MS-NN significantly improves tracking accuracy, decreases steering oscillations, and is physically interpretable. Moreover, online velocity replanning improves lap times by compensating for execution errors, and enables the vehicle to safely reach higher speeds and accelerations. To support future research, our code, datasets, videos and results are publicly available at \url{https://roboracer-benchmark.github.io/planning_control_benchmark/}.
    \end{abstract}

	\section{Introduction} \label{sec:introduction}
    \begin{figure*}[t]
        \centering
        \includegraphics[width=0.96\linewidth]{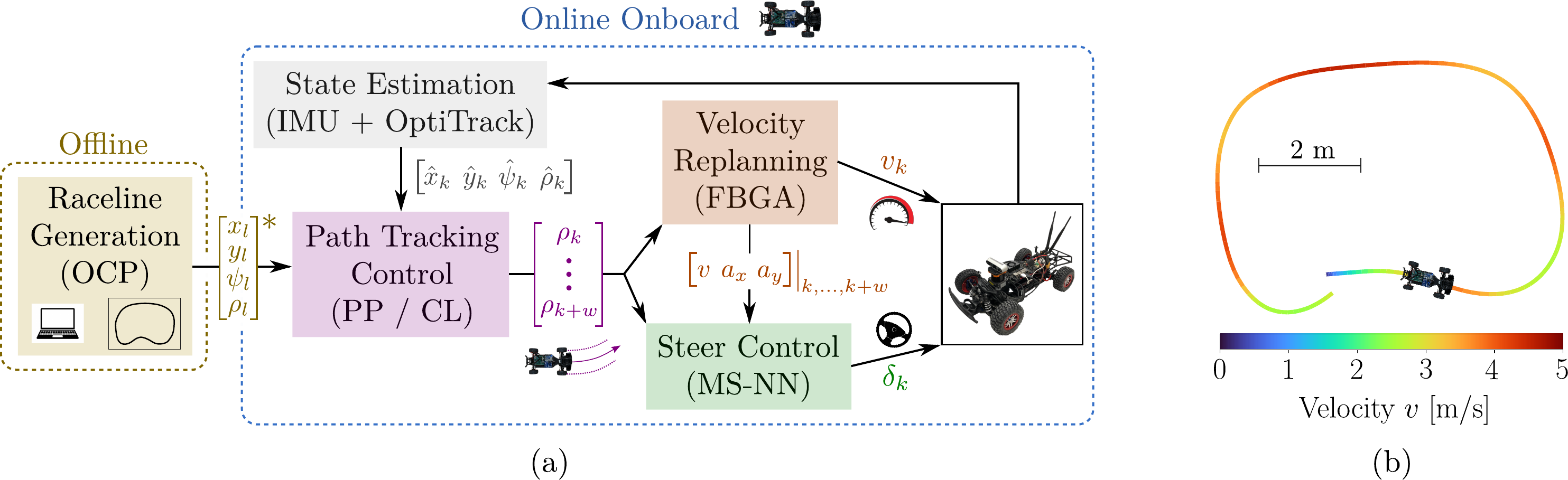}
        \caption{(a) Overview of our trajectory planning and control framework, and (b) real-world results obtained by deploying the framework on our RoboRacer.}
        \label{fig:framework_overview}
    \end{figure*}
    Despite recent advances in autonomous driving, planning and executing high-acceleration maneuvers in emergency-like scenarios remains an open challenge. To foster research in this direction, autonomous racing has emerged as a safe testbed for advanced planning and control strategies \cite{betz2022autonomous}, where vehicles operate at the handling limits to achieve competitive lap times.

    Modular software stacks with perception-planning-control (PPC) are widely adopted in autonomous driving and racing \cite{betz2022autonomous}, where each module is designed for a specific role. Conversely, end-to-end (E2E) approaches learn a mapping from sensor inputs to control commands using neural networks (NNs). Although reinforcement learning (RL) achieved superhuman results in simulation \cite{Wurman2022}, fully-E2E methods still lack generalization, and reliable real-world transfer remains challenging. In the following, we review modular PPC approaches, including learning-based components. 
    \subsubsection{Global Trajectory Generation}
    In autonomous racing, global time-optimal racelines over a full circuit are typically computed by solving Optimal Control Problems (OCPs), using either point-mass models \cite{Rowold2023,Piccinini2025ICRA} or more complex nonlinear single- and double-track models \cite{subosits2021impacts,Christ2021,Piccinini_2024_how_optimal}. Minimum-curvature racelines have also been proposed \cite{Heilmeier2020}, though they can significantly differ from time-optimal ones. Given its computational cost, raceline optimization is typically performed offline over the full track before execution.
    \subsubsection{Lateral Trajectory Tracking}
    Lateral tracking controllers \cite{Kebbati2023} generate local paths and steering commands to follow upstream reference trajectories. Geometric methods, such as Pure Pursuit (PP) \cite{coulter1992implementation} and Stanley variants \cite{AbdElmoniem2020}, are popular for their simplicity and real-time capability, but fail to handle the nonlinear vehicle dynamics near the limits. Model-based approaches, like Model Predictive Control (MPC), explicitly incorporate nonlinear dynamics and constraints \cite{Vazquez2020,Jahncke2026_diff_MPC}. However, performing at the dynamic limits requires complex vehicle models, increasing computational load and limiting real-time use on embedded platforms.
    Recently, learning-based methods have been introduced to augment geometric and model-based controllers. In \cite{Ghignone2025}, an RL agent provides a steering correction on top of a PP controller, while \cite{Kabzan2019} enhanced MPC by modeling single-track residuals with Gaussian Processes.
    NNs have also been used to learn the vehicle dynamics and generate steering commands. General-purpose recurrent and convolutional NNs were adopted in \cite{Devineau2018,piccinini2023predictive,Gottschalk2024}, while \cite{dalio2020mental,Piccinini2023_physics_driven,piccinini2025model,piccinini2025road} proposed Model-Structured NNs (MS-NNs) with architectures inspired by vehicle dynamics laws. Although MS-NNs show improved generalization and sample efficiency over generic NNs, they have not yet been validated on real vehicles at high accelerations, and their closed-loop operation is unexplored.
    \subsubsection{Online Velocity Planning}
    %
    In racing, online speed replanning is crucial to compensate for execution errors and varying track conditions.
    In \cite{frego2017semi}, a forward-backward (FB) scheme was proposed for online minimum-time speed planning, yet limited to box constraints on longitudinal and lateral accelerations. This approach was later extended to handle super-elliptic \cite{Langmann2025} and general \GGV{} constraints \cite{Piazza2025} (FBGA), capturing the nonlinear coupling between longitudinal and lateral accelerations near the limits.
    Our framework uses the FBGA method of \cite{Piazza2025}, which has never been deployed on physical systems.
    %
    \subsubsection{Critical Literature Summary} 
    To our knowledge, the community lacks open-source benchmarks to evaluate new planning and control methods in high-acceleration, race-like scenarios, with real-world validation on open platforms, open datasets, and code.
    Moreover, although MS-NN steering control and online time-optimal velocity replanning show promising simulation results, their closed-loop performance in PPC stacks for real-world racing is largely unexplored.
    
    To address these gaps, our contributions are as follows:
    \begin{itemize}
        \item We benchmark new and existing methods for time-optimal planning and control near the handling limits, including geometric controllers, MS-NN steering controllers, and an online velocity planner. These methods have never been used in a unified closed-loop stack for real-world autonomous racing.
        \item We present a new MS-NN that learns the nonlinear coupled longitudinal-lateral dynamics for steering control at high accelerations, while preserving interpretability.
        \item Through several ablations, we analyze the performance of single modules and their combinations, quantifying the impact of the MS-NN and online velocity replanning on trajectory planning, tracking, and lap-time.  
        \item We conduct our experiments on a 1:10-scale autonomous racing platform, and we publicly release all software modules, datasets, and results.
    \end{itemize}
    

    
	\section{Trajectory Planning and Control Framework}
	\label{sec:designandimplementation}
    \subsection{Framework Overview}
    Our trajectory planning and control framework is shown in Fig.~\ref{fig:framework_overview}a. From the track boundaries of a given circuit, we generate offline a time-optimal raceline using the open-source tool\footnote{\href{https://github.com/TUMFTM/global\_racetrajectory\_optimization}{https://github.com/TUMFTM/global\_racetrajectory\_optimization}} of \cite{Christ2021}, based on a point-mass model with user-defined acceleration and velocity constraints.
    Online, a path-tracking controller computes future path curvature references $\boldsymbol{\rho} = [\rho_k, \dots, \rho_{k+w}]$ from the current vehicle state, to follow the raceline. These are passed to a velocity replanning module, which computes time-optimal velocity and longitudinal-lateral acceleration profiles $\{\boldsymbol{v}, \boldsymbol{a_x}, \boldsymbol{a_y}\}$ over a planning horizon via a forward-backward optimizer (FBGA) \cite{Piazza2025}, accounting for the vehicle's nonlinear constraints on accelerations and velocity. The velocity profile is tracked by the motor controller, while steering commands $\delta$ are generated by a new MS-NN using curvature, velocity, and acceleration references. The estimated vehicle state closes the loop.
    The full framework runs in real time (at 150 Hz) on our 1:10-scale RoboRacer (Sec.~\ref{sec:experimental_setup}). 
    \subsection{Path Tracking Controllers} \label{sec:path_tracking_controllers}
    We compare the following two path tracking controllers.
    
    \subsubsection{Pure Pursuit (PP)} This algorithm connects the current vehicle position and heading $[\hat{x}, \hat{y}, \hat{\psi}]$ to a dynamically selected look-ahead point $[x^*_{l}, y^*_{l}]$ on the raceline path using a circular arc \cite{coulter1992implementation}. The (constant) arc path curvature is given by:
    \begin{equation}
        \rho = 2 \, \sin(\lambda) \, / \, l_d \; ,
    \end{equation}
    %
    %
    where $l_d$ the tunable look-ahead distance, and $\lambda$ the angle between the vehicle's longitudinal axis and the line connecting the vehicle pose to the look-ahead point.
    %
    %
    \subsubsection{Clothoid-based (CL)} 
\begin{figure}[]
        \centering
        \includegraphics[width=0.33\textwidth]{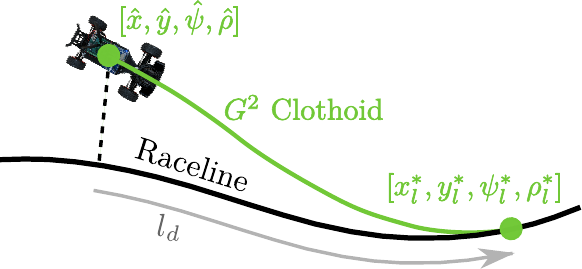}
        \caption{Clothoid-based (CL) path tracking controller: we connect the current vehicle pose and curvature to a look-ahead ($l_d$) raceline point using a $G^2$ clothoid, which enforces continuity of position, heading, and curvature.}
        \label{fig:cl_schematic}
    \end{figure}

In this controller, we propose to connect the current vehicle pose and curvature $[\hat{x}, \hat{y}, \hat{\psi}, \hat{\rho}]$ to a look-ahead raceline point $[x^*_{l}, y^*_{l}, \psi^*_{l}, \rho^*_{l}]$ using a $G^2$ clothoid \cite{Bertolazzi2018} (Fig. \ref{fig:cl_schematic}). This curve enforces continuity of position, heading, and curvature, which is beneficial for high-acceleration maneuvers.
    We generate the clothoid and compute its curvature $\rho(s)$ numerically using the open-source tool\footnote{\href{https://github.com/ebertolazzi/Clothoids}{https://github.com/ebertolazzi/Clothoids}.} of \cite{Bertolazzi2018}, where $s$ denotes the curvilinear abscissa along the curve. 
    
    In our framework (Fig.~\ref{fig:framework_overview}), the PP and CL controllers generate the curvature reference $\boldsymbol{\rho}$ for the MS-NN, which outputs the steering command $\delta$. However, our ablations also evaluate PP and CL without the MS-NN, computing $\delta(s)$ along the curve directly from $\rho(s)$ via a kinematic bicycle model:
    \begin{equation}
        \label{eq:kine_steering}
        \delta(s) = \arctan\big(\rho(s) \, L\big) \; ,
    \end{equation}
    where $L$ is the vehicle's wheelbase.
    \subsection{Online Time-Optimal Velocity Replanning} \label{sec:velocity_replanning}
    Given the future curvature profile $\rho(s)$ from the path-tracking controller, our velocity replanner computes online time-optimal velocity and acceleration profiles ${v(s), a_x(s), a_y(s)}$ along the path. We adapt the open-source forward-backward method (FBGA) of \cite{Piazza2025}, based on a point-mass model with constraints on ${a_x, a_y}$ and $v$ defined by the \GGV{} diagram (Fig.\ref{fig:limit_ggv}). The same \GGV{} is used for both online replanning and offline raceline generation (Sec.\ref{sec:raceline_generation}), thus ensuring consistency between the two modules.
    %
    %
    \subsection{Steering Control: Model-Structured Neural Network} \label{sec:steering_control_msnn}
    \begin{figure}[]
        \centering
        \includegraphics[width=0.38\textwidth]{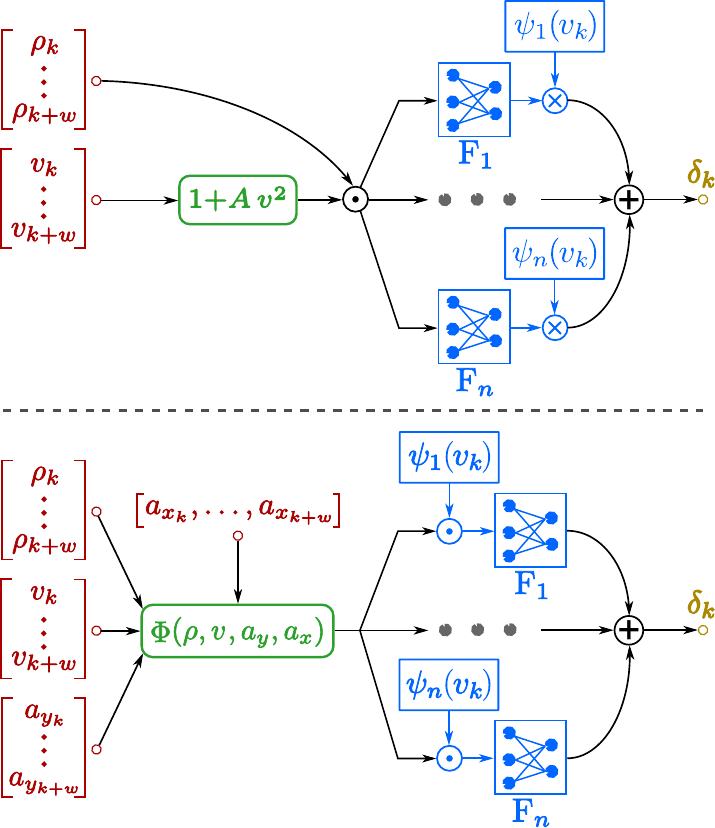}
        \caption{MS-NN steer controller of \cite{dalio2020mental} (top), and our extension (bottom).} 
        \label{fig:ms_nn}
    \end{figure}
    We introduce a new Model-Structured Neural Network (MS-NN) to generate steering commands for the vehicle's actuator. As shown in Fig. \ref{fig:framework_overview}a and Fig. \ref{fig:ms_nn} (bottom), our MS-NN computes the steering angle $\delta_k$ at time step $k$ from the following future reference signals:
    \begin{subequations}\label{eq_vectors_future_inputs}
    \begin{empheq}[left=\empheqlbrace]{alignat=2}
        &\text{Path curvature:}         & \;\;\boldsymbol{\rho}_k &= [ \rho_{k}, \dots, \rho_{k+w} ] \label{eq_future_rho} \\
        &\text{Longitudinal velocity:}  & \;\; \boldsymbol{v}_k &= [ v_{k}, \dots, v_{k+w} ] \label{eq_future_v} \\
        &\text{Lateral acceleration:}   & \;\; \boldsymbol{a_y}_k &= [a_{y_k}, \dots, a_{y_{k+w}} ] \label{eq_future_ay} \\
        &\text{Longitud. acceleration:} & \;\; \boldsymbol{a_x}_k &= [ a_{x_k}, \dots, a_{x_{k+w}} ] \label{eq_future_ax}
    \end{empheq}
    \end{subequations}    
    where $w$ is the future window length with sampling time $T_s$, both tunable. The curvature $\boldsymbol{\rho}_k$ is given by the path tracking controller; the velocity $\boldsymbol{v}_k$ and acceleration profiles $\{\boldsymbol{a_y}_k, \boldsymbol{a_x}_k\}$ are computed by the online velocity planner.

    We formulate our MS-NN as follows (Fig. \ref{fig:ms_nn}, bottom):
    \begin{equation}
        \delta_{k} = \sum_{i=1}^{n} \Big( \underbrace{\mathbf{\Phi}(\boldsymbol{\rho}_k, \boldsymbol{v}_k, \boldsymbol{a_y}_k, \boldsymbol{a_x}_k)}_{\text{quasi steady-state}} \odot \, \overbrace{\boldsymbol{\psi}_{i}(\boldsymbol{v}_k) \Big) \cdot \begin{bmatrix} F_{i_1} \\    \vdots \\ F_{i_{w+1}} \end{bmatrix}}^{\text{transient}} \; ,
    \label{eq:MS-NN-base}
    \end{equation}
    where $\odot$ is the element-wise vector product. The first term $\mathbf{\Phi}(\cdot)$ learns the \textit{quasi steady-state} mapping from the future reference signals to the steering command, while the second term learns velocity-dependent \textit{transient} effects.
    
    \textit{Quasi steady-state:} We design the function $\mathbf{\Phi}(\cdot)$ to capture the shape of the handling diagram (HD) \cite{Guiggiani2018}, which describes the deviation of the actual steering command $\delta$ from the perfect kinematic steering angle $\rho L$ (Eq. \eqref{eq_quasi_steady_state}):
    \begin{subnumcases}{\label{eq_quasi_steady_state_full}}
    Q(v,a_x) = q_0 + q_v v + q_x a_x\\ 
    S(v,a_x) = s_0 + s_v v + s_x a_x\\
    \delta - \rho L = \Phi(\rho, v, a_y, a_x) = Q(v,a_x) \, a_y + S(v,a_x) \,a_y^2 \label{eq_quasi_steady_state}
    \end{subnumcases}
    where \eqref{eq_quasi_steady_state} is a second-order relation in $a_y$, with coefficients $Q$ and $S$ linear in $v$ and $a_x$. This structure captures the nonlinear HD shape and its dependence on velocity and longitudinal acceleration, while remaining interpretable. The parameters $\{q_0, q_v, q_x, s_0, s_v, s_x\}$ are learnable.
    
    \textit{Transient:} We use fully connected layers $\boldsymbol{F}_i$, $i \in \{1,\dots,n\}$ (blue blocks in Fig.~\ref{fig:ms_nn}), to learn linear filters that combine the future reference entries of $\boldsymbol{\Phi}$ into $n$ scalar features. Following the physics insight that the vehicle transient dynamics are velocity-dependent \cite{Guiggiani2018}, we adopt $n$ local $\boldsymbol{F}_i$ layers, each activated by a velocity-dependent gating function $\boldsymbol{\psi}_i(\boldsymbol{v})$, which allows the network to learn different transient effects in different velocity ranges. Like in \cite{dalio2020mental}, we design the gating functions as triangular neuro-fuzzy functions of $v$, with $\sum_{i=1}^{n} \boldsymbol{\psi}_i(\boldsymbol{v}) = 1, \forall \boldsymbol{v}$.
    The final steering command is obtained by summing the contributions of all $n$ features.

    The training, evaluation, hyperparameters tuning, and interpretation of the proposed MS-NN are detailed in Sec.~\ref{sec:msnn_training_evaluation}.
    \subsubsection{Baseline MS-NN} \label{sec:baseline_msnn}
    The proposed MS-NN extends the baseline in \cite[Sec. III-B]{dalio2020mental} (Fig. \ref{fig:ms_nn}, top), conceived for low-acceleration maneuvers. The baseline assumes a linear HD with a constant learnable slope $A$ (green block in Fig. \ref{fig:ms_nn}, top) and ignores the dependence on $v$ and $a_x$. Its transient component also differs slightly: the gating functions $\psi_i(v_k)$ are driven only by the current velocity $v_k$, rather than the full future profile $\boldsymbol{v}_k$, limiting the learning of velocity-dependent transients. In contrast, our MS-NN captures the nonlinear coupled longitudinal-lateral steering behavior, while preserving a compact, interpretable structure. A comparison with the baseline is provided in Sec.~\ref{sec:msnn_training_evaluation}.
\section{Experimental Results}
All results in this paper are obtained through \textbf{real-world experiments} with the following vehicle platform and setup. 
\begin{figure}
	\centering
    \subfloat[][RoboRacer vehicle.]{\includegraphics[width=0.505\linewidth]{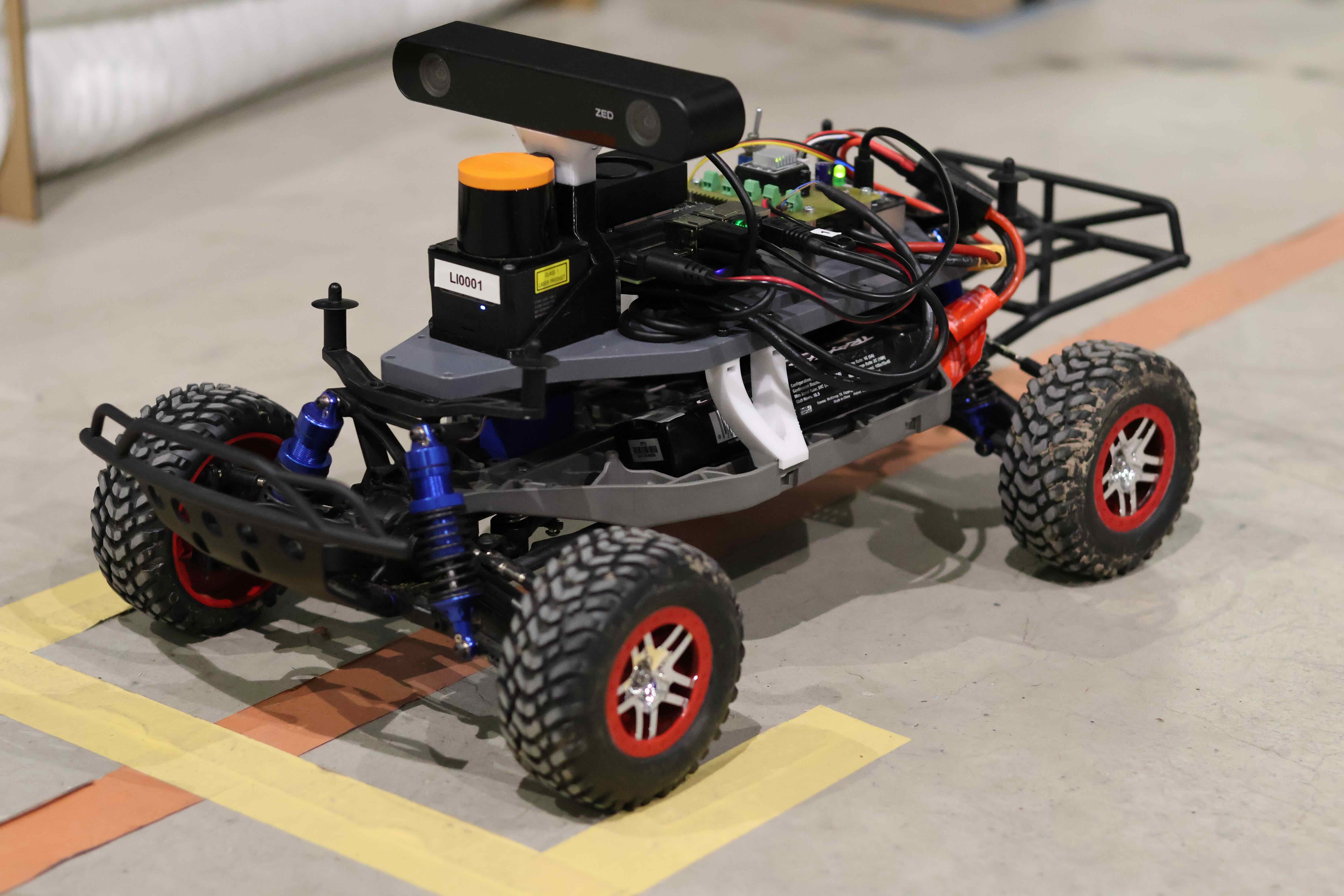}\label{fig:Vehicle}}\\
	\subfloat[][Track A layout.]{\includegraphics[width=0.48\linewidth]{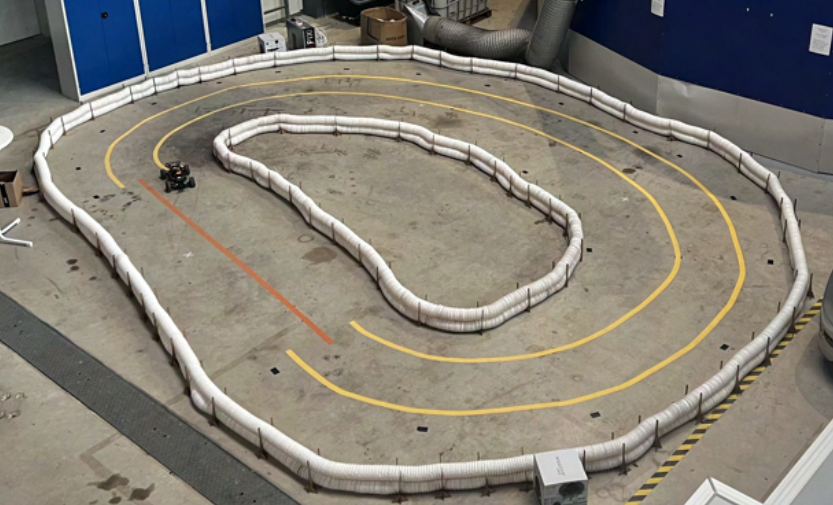}\label{fig:Track_A}}\hfill%
    \subfloat[][Track B layout.]{\includegraphics[width=0.48\linewidth]{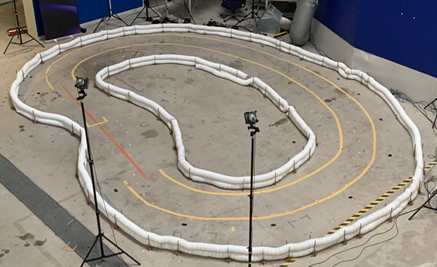}\label{fig:Track_B}}
	\caption{RoboRacer vehicle and track layouts used in our experiments.}
	\label{fig_experimental_setup}
\end{figure}

\subsection{Experimental Setup}
\subsubsection{RoboRacer Platform} \label{sec:experimental_setup}
We validate the proposed algorithms on a 1:10-scale robotic vehicle, the RoboRacer\footnote{\href{https://roboracer.ai}{https://roboracer.ai}} (Fig. \ref{fig:Vehicle}). The RoboRacer is a widely used, low-cost platform for robotics research and autonomous racing, enabling rapid prototyping of autonomous driving algorithms. The vehicle features an all-wheel-drive electric powertrain powered by a 300 W Velineon 3351R/3500 brushless DC motor, actuated via a Veddar electronic speed controller (VESC).

We execute our motion planning and control algorithms on an onboard NVIDIA Jetson Orin Nano, running a Ubuntu-based operating system. Our autonomous software stack is implemented using the ROS 2 middleware.
In our experiments, we feed the requested steering angles to the steering actuator, while the desired vehicle speed is tracked by the VESC motor controller. We measure the yaw rate, the longitudinal and lateral accelerations through an onboard IMU. The vehicle pose and velocity are estimated by an external motion capture system (OptiTrack), which provides accurate high-frequency localization for performance evaluation.
\subsubsection{Test Tracks}
We evaluate the proposed methods on two indoor circuits, referred to as \textit{Track A} and \textit{Track B}, designed with increasing geometric complexity.
Track A (Fig.~\ref{fig:Track_A}) is a near-oval baseline layout, approximately $18$ m long and $1.6-2.1$ m wide. 
Track B (Fig.~\ref{fig:Track_B}), extends Track A by introducing an additional turn, increasing curvature variability and inducing more frequent and asymmetric steering actions. 
%
\subsection{Racelines Generation and \GGV{} Diagrams} \label{sec:raceline_generation} 
For both tracks, we generate time-optimal racelines following \cite{Christ2021}, which solves a point-mass OCP with user-defined \GGV{} diagram constraints. The \GGV{} diagram defines the reachable longitudinal and lateral accelerations $\{a_x,a_y\}$ as a function of the vehicle speed $v$ (Fig. \ref{fig:limit_ggv}). We use three \GGV{} envelopes to generate racelines of varying aggressiveness, as detailed below.
\subsubsection{Cautious \GGV{}} \label{sec:cautious_ggv}
This envelope is designed to be conservative, ensuring that the generated raceline can be tracked without the online velocity replanning module\footnote{In the absence of online velocity replanning, the vehicle tracks the velocity profile of the offline-generated raceline without real-time changes.}. This allows us to assess the performance of the geometric controllers (PP and CL) and the MS-NN steering module in isolation. We use an elliptical \GG{} diagram with acceleration limits $a_{y,\max} = 3.0~\mathrm{m/s^2}$, $a_{x,\max} = 1~\mathrm{m/s^2}$, and $a_{x,\min} = -1~\mathrm{m/s^2}$, which we empirically determine to ensure stable operation of the geometric controllers on our platform.
\subsubsection{Nominal \GGV{}} \label{sec:nominal_ggv}
This envelope (Fig.~\ref{fig:limit_ggv}, transparent dashed line) aims to push the coupling of the PP / CL tracking controllers and the MS-NN to the limits, in the absence of the online velocity replanning module. We identify this \GGV{} envelope iteratively: starting from conservative limits, the acceleration bounds are increased until closed-loop instability occurs, defined by excessive path or velocity errors, track bounds violations, or loss of control.
\subsubsection{Extended \GGV{}} \label{sec:extended_ggv}
\begin{figure}[]
    \centering
    \includegraphics[width=0.96\linewidth]{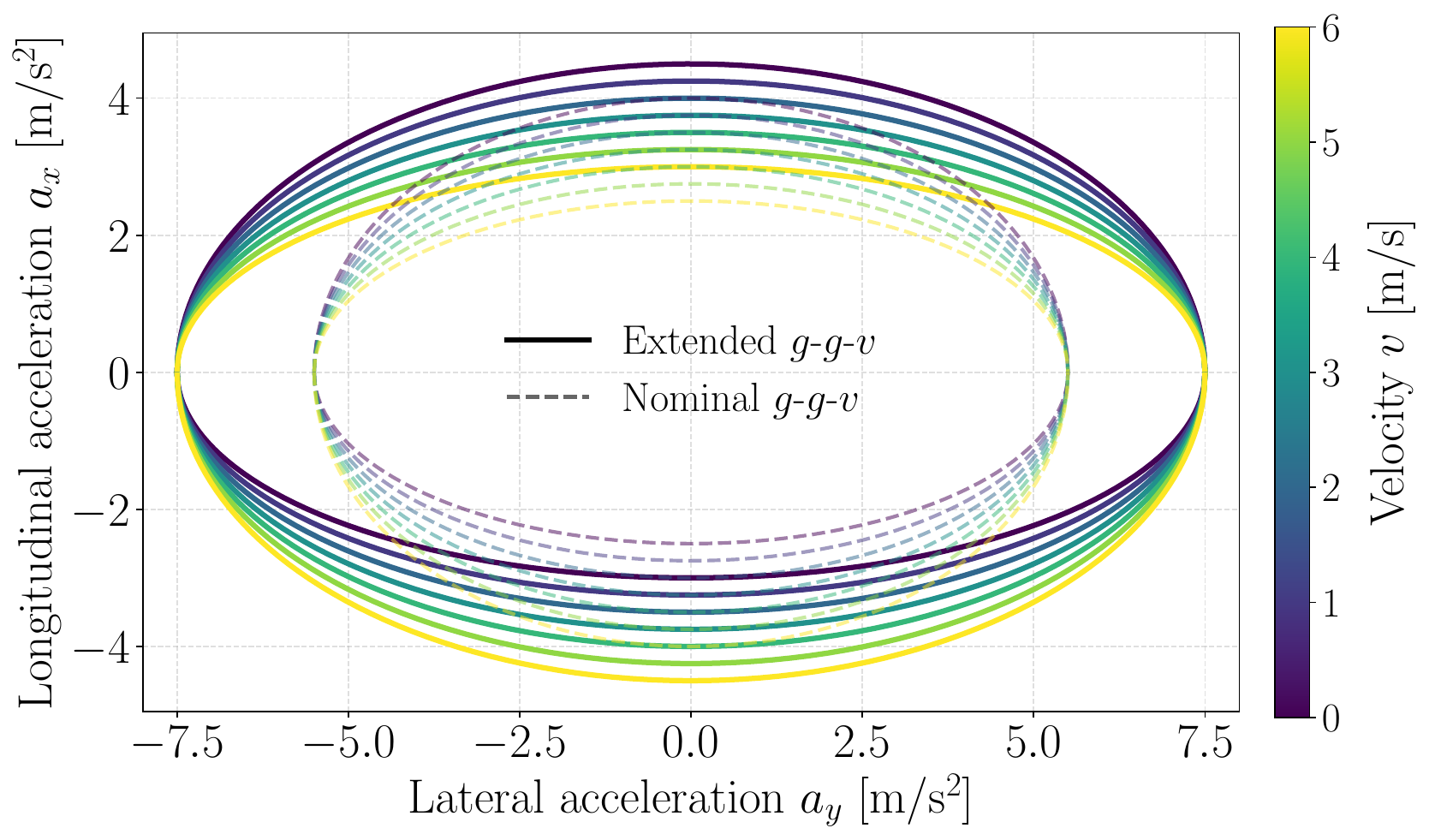}
    \vspace{-0.3cm}
    \caption{Velocity-dependent acceleration limits (\GGV{} diagrams) used as constraints for offline raceline generation and online velocity replanning.}
    \label{fig:limit_ggv}
\end{figure}

This envelope aims to bring our full framework (including online velocity replanning) to its limits. Starting from the nominal \GGV{}, we iteratively expand it until our closed-loop feasibility indicators are violated, as defined in the previous paragraph. As shown in Fig.~\ref{fig:limit_ggv}, the extended \GGV{} allows significantly higher accelerations, enabled by online velocity replanning, which computes time-optimal profiles while preserving dynamic feasibility.
%
%
%
%
\subsection{MS-NN Training, Evaluation and Interpretation} \label{sec:msnn_training_evaluation}
\subsubsection{Training}
The MS-NN steering controller is trained via supervised learning, on data collected from closed-loop experiments with the vehicle following the nominal raceline on Track B using the pure pursuit controller. Our loss function is the Root Mean Square Error (RMSE) between predicted and measured steering angles $\delta$.
The training and validation sets comprise approximately $18$ and $6$ seconds of driving data, respectively (Fig. \ref{fig:msnn_training_validation}), highlighting the sample efficiency of the proposed architecture.
Training uses Adam with a learning rate of $10^{-4}$, batch size 200, and up to 6000 epochs, with early stopping (patience 400) to mitigate overfitting.
\subsubsection{Evaluation}
\begin{figure}[]
    \centering
    \includegraphics[width=0.94\linewidth]{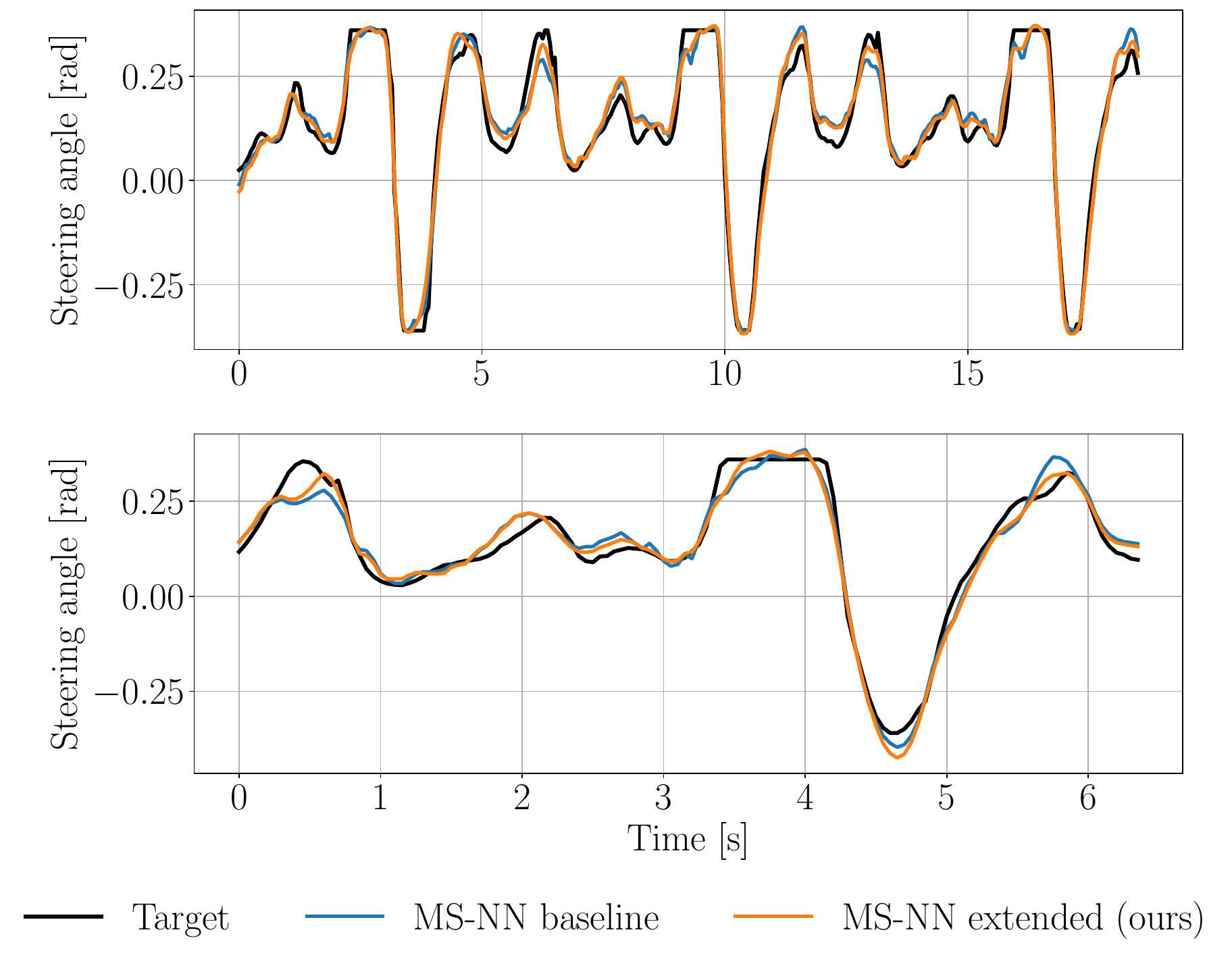}
    \vspace{-0.5cm}
    \caption{Measured and predicted steering angles on the training and validation sets: our MS-NN outperforms the baseline \cite{dalio2020mental}.}
    \label{fig:msnn_training_validation}
    \vspace{-0.3cm}
\end{figure}

Performance is evaluated using the RMSE of the steering prediction error and the Fraction of Variance Unexplained (FVU), which measures the proportion of ground-truth variance not captured by the model (lower is better). The proposed MS-NN outperforms the baseline of \cite{dalio2020mental} (Sec. \ref{sec:baseline_msnn}) by 8.4\% RMSE and 20.9\% FVU on the training set, and by 10.6\% RMSE and 22.7\% FVU on the validation set (Table~\ref{tab:msnn_comparison}). Fig.~\ref{fig:msnn_training_validation} further confirms the improvement, showing more accurate local prediction of steering peaks and transient dynamics. The consistent gains across datasets indicate improved generalization and a more accurate learning of the underlying steering dynamics.
\begin{table}[]
    \centering
    \caption{Comparing the baseline and proposed MS-NN variants.}
    \label{tab:msnn_comparison}
    \begin{tabular}{l c c c}
    \toprule
    \textbf{Dataset} & \textbf{MS-NN} & \textbf{RMSE} $\downarrow$ [m] & \textbf{FVU} $\downarrow$ [-] \\
    \midrule
    \multirow{2}{*}{Training} & 
    Baseline \cite{dalio2020mental}     
    & 0.0357 & 0.0454 \\
    & Extended (ours) & \textbf{0.0327} & \textbf{0.0359} \\
    \midrule 
    \multirow{2}{*}{Validation} &
    Baseline \cite{dalio2020mental}
    & 0.0365 & 0.0423 \\
    & Extended (ours) & \textbf{0.0326} & \textbf{0.0327} \\
    \bottomrule
    \end{tabular}
\end{table}

\subsubsection{Hyperparameters Tuning}
Hyperparameters are tuned via a grid search, selecting the configuration that minimizes the Akaike Information Criterion (AIC) \cite{Bozdogan1987} on the validation set. The AIC balances accuracy and model complexity (number of learnable parameters), with lower values indicating better sample efficiency.
Table~\ref{tab:msnn_sensitivity} reports results for different numbers of velocity channels $n$ and input window lengths $(w+1)$. The best configuration is $n = 1$ and $w = 9$, yielding the lowest AIC. With a discretization step $T_s = 0.05$ s, this corresponds to input windows spanning $(w+1)\, T_s = 0.5$ s, sufficient to capture the dynamics while keeping the model compact. 
\begin{table}[]
    \centering
    \caption{Hyperparameter tuning of our MS-NN, based on the AIC metric on the validation set. The best configuration is highlighted.}
    \label{tab:msnn_sensitivity}
    \begin{tabular}{c c c}
    \toprule
    \textbf{N. channels $n$} & \textbf{N. future samples $w$} & \textbf{AIC} $\downarrow$ [-] \\
    \midrule
    \multirow{2}{*}{3} & 14 & -411.0 \\
                        & 9 & -457.5 \\
    \midrule
    \multirow{2}{*}{2} & 14 & -426.8 \\
                        & 9 & \underline{-469.3} \\
    \midrule
    \multirow{2}{*}{1} & 14 & -450.5 \\
                        & 9 & \textbf{-486.8} \\
    \bottomrule
    \end{tabular}
\end{table}

\subsubsection{Implementation}
We formulate and train our MS-NN using \texttt{nnodely}\footnote{\href{https://github.com/tonegas/nnodely.git}{https://github.com/tonegas/nnodely.git}}, an open-source framework for structured and physics-encoded neural networks. The library offers a high-level interface to define custom architectures, input transformations, and loss functions, while providing efficient training.
\subsubsection{Interpreting the Learned Weights}
\begin{figure}[]
    \centering
    \includegraphics[width=0.93\linewidth]{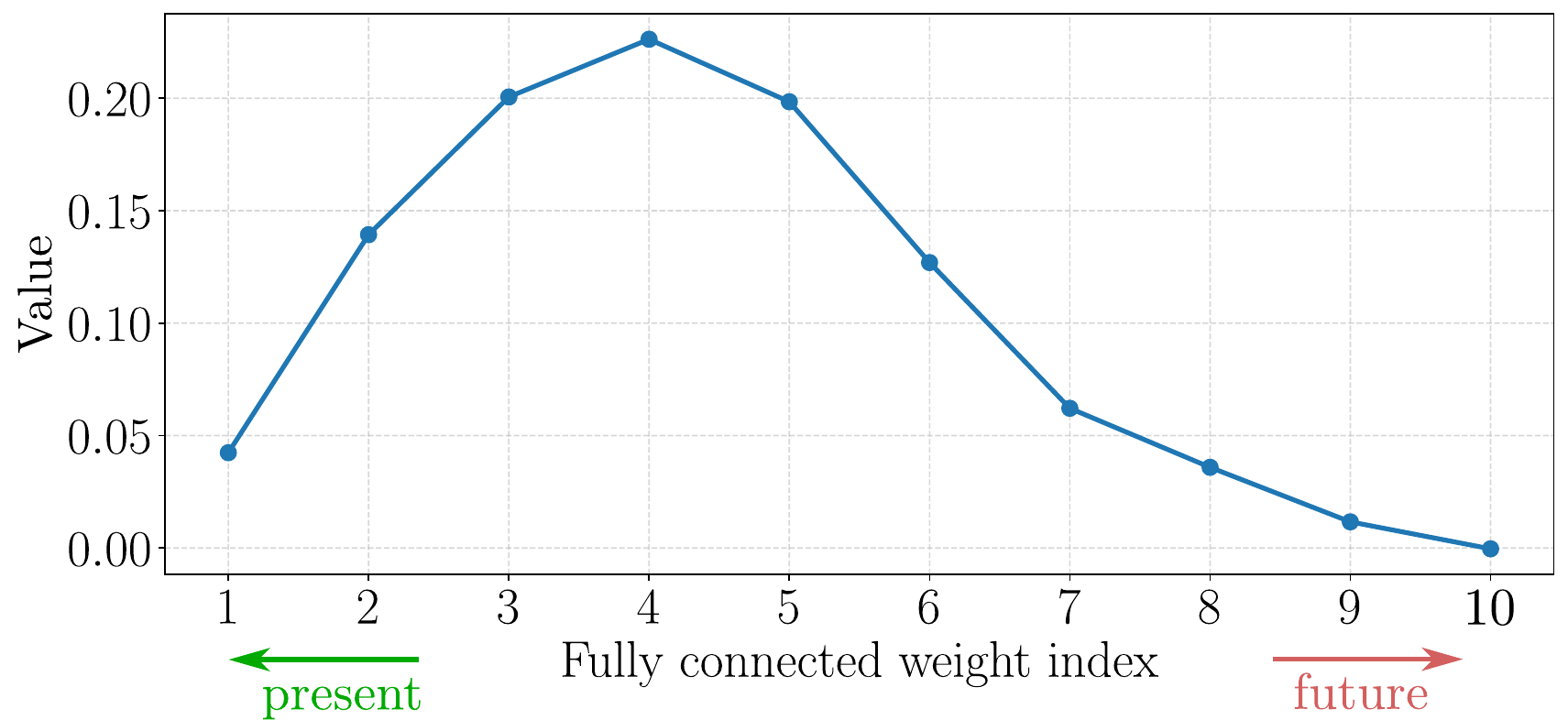}
    \caption{Interpreting the trained weights of the fully connected layer $\boldsymbol{F}$ in our MS-NN. The peak at the 4$^{\mathrm{th}}$ index (corresponding to $0.2$ s ahead) indicates that future references around this time have the strongest influence on the current steering command, revealing a delay in the steering response.}
    \label{fig:FIR_weights_comparison}
\end{figure}

The compact MS-NN structure enables direct interpretation of the learned weights and the underlying steering dynamics.
For example, we consider the fully connected layer $\boldsymbol{F}$ (blue blocks in Fig.~\ref{fig:ms_nn}, with $n=1$), which acts as a linear filter combining the $10$ ($= w+1$) future reference entries of $\mathbf{\Phi}$ into a scalar feature.
Fig.~\ref{fig:FIR_weights_comparison} shows that the learned weights of $\boldsymbol{F}$ peak at the 4$^{\mathrm{th}}$ index, corresponding to $0.2$ s ($4\,T_s$) ahead. This indicates that future references around $0.2$ s ahead most strongly influence the current steering command, revealing a delay in the steering dynamics.
Hence, these weights reflect the vehicle's dynamic response, highlighting delays and characteristic time scales.
%
%

\subsection{Online Trajectory Tracking without Velocity Replanning}
\begin{table}[]
\centering
\caption{Trajectory tracking on track A with the cautious raceline and without velocity replanning. Our MS-NN improves path tracking, lap times, and steering smoothness.}
\label{tab:trackA_steering_metrics}
\begin{tabular}{l|c|c|c|c}
\toprule
\multirow{2}{*}{Controller} & \multicolumn{2}{c|}{Path Tracking [cm]} & Lap Time & RMS Steering Rate \\
& Mean & Max &  [s] & [rad/s] \\
\midrule
CL         & 5.17 & 9.20 & 7.065 & 0.2146 \\
PP         & 5.13 & 8.23 & 7.039 & 0.1323 \\
CL + MS-NN & \underline{4.09} & \underline{8.20} & \underline{6.923} & \underline{0.0917} \\
\textbf{PP + MS-NN} & \textbf{4.04} & \textbf{7.92} & \textbf{6.919} & \textbf{0.0841} \\
\bottomrule
\end{tabular}
\end{table}
This section evaluates raceline tracking without velocity replanning, to isolate the effects of the PP and CL controllers (Sec. \ref{sec:path_tracking_controllers}) and our MS-NN steering module (Sec. \ref{sec:steering_control_msnn}). The local velocity reference is taken directly from the offline-generated raceline. Experiments are conducted on Track A (Fig.~\ref{fig:Track_A}) using the cautious raceline (Sec. \ref{sec:cautious_ggv}) that does not require velocity replanning to be tracked. For PP and CL, we use speed-dependent look-ahead distances, tuned to minimize tracking errors: $0.8$-$1.4$ m for PP and $1.0$-$1.4$ m for CL.
%
%
\subsubsection{PP vs. CL without MS-NN} 
When the MS-NN is excluded from the control loop, and PP / CL compute the steering angle directly via Eq. \eqref{eq:kine_steering}, PP outperforms CL (Table~\ref{tab:trackA_steering_metrics} and Fig.~\ref{fig:trackA_boxplot}), achieving 10.5\% lower peak tracking error, better steering smoothness (Fig.~\ref{fig:trackA_steering}), and slightly improved lap time.

\subsubsection{Impact of our MS-NN}
\begin{figure}[]
    \centering
    \includegraphics[width=0.9\linewidth]{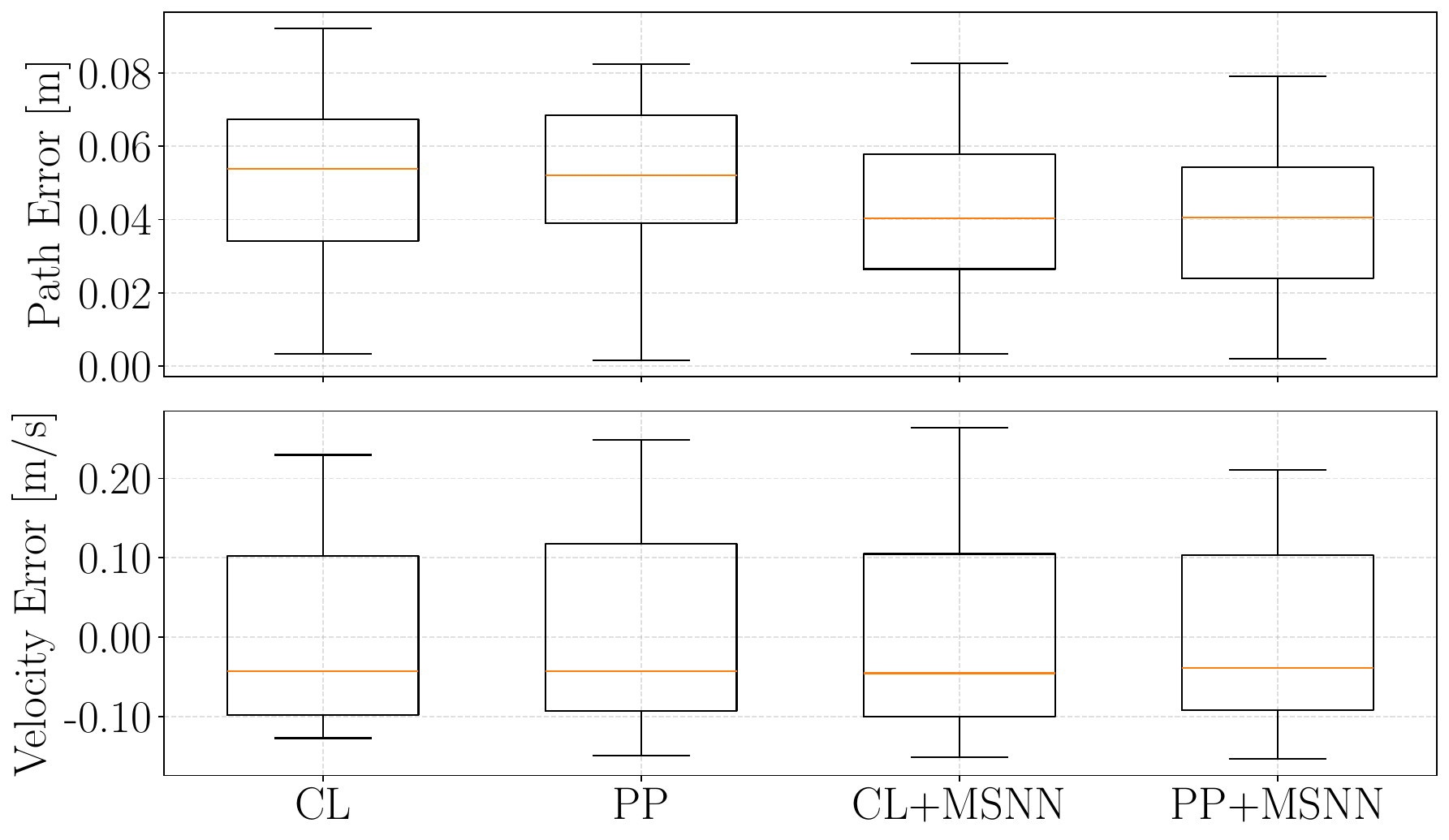}
    \caption{Path and velocity deviations from the raceline on track A, with the cautious raceline and without velocity replanning (as in Table~\ref{tab:trackA_steering_metrics}). Our MS-NN reduces median and maximum deviations.}
    \label{fig:trackA_boxplot}
\end{figure}

Coupling PP and CL with our MS-NN for steering control further improves trajectory tracking and lap times (Table~\ref{tab:trackA_steering_metrics}, Fig.~\ref{fig:trackA_boxplot}). The best configuration, PP + MS-NN, reduces mean path tracking error by 21.2\% and lap time by 1.7\% relative to PP alone. Similarly, CL + MS-NN improves peak tracking error by 10.8\% compared to CL. 

\begin{figure}[]
    \centering
    \includegraphics[width=0.92\linewidth]{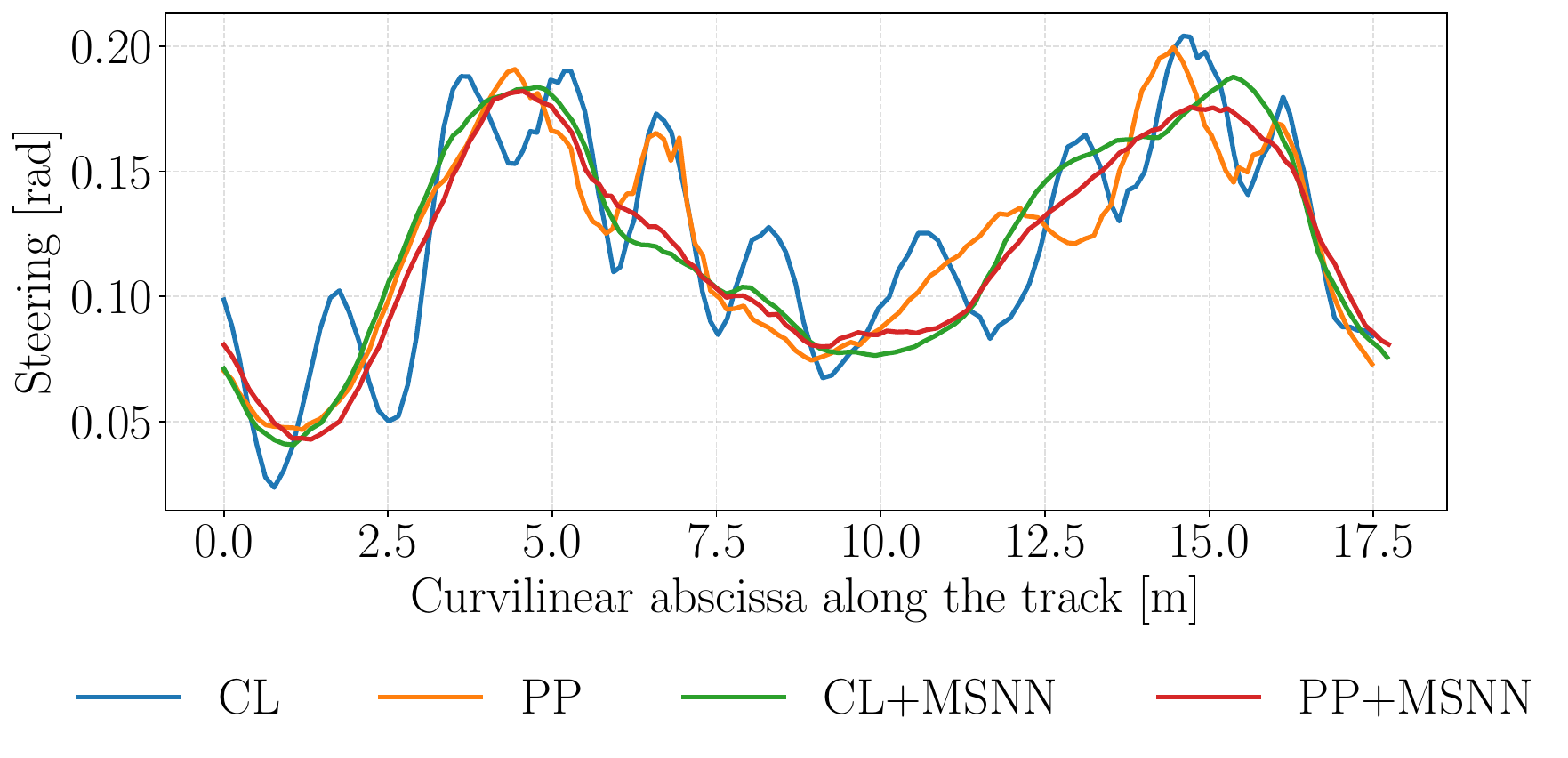}
    \caption{Steering angles generated by the four controller combinations on track~A, with the cautious raceline and without velocity replanning. Coupling our MS-NN with PP / CL significantly reduces steering oscillations (Table~\ref{tab:trackA_steering_metrics}).}
    \label{fig:trackA_steering}
\end{figure}

Moreover, our MS-NN markedly reduces steering oscillations (Table~\ref{tab:trackA_steering_metrics}, Fig.~\ref{fig:trackA_steering}): PP + MS-NN lowers the RMS steering rate by 36.4\% relative to PP, while CL + MS-NN reaches a 57.2\% reduction compared to CL. Improved steering smoothness enhances the vehicle stability near the handling limits by limiting aggressive corrective inputs that may induce a loss of control. Finally, velocity tracking remains comparable across controllers (Fig.~\ref{fig:trackA_boxplot}), since all follow the same offline velocity profile without online replanning. Notably, PP + MS-NN shows a slight reduction in peak velocity error, attributable to improved path tracking and reduced steering oscillations, which indirectly enhance velocity tracking.

Overall, these results demonstrate that the MS-NN effectively captures unmodeled dynamics and compensates for the limitations of the geometric PP and CL controllers alone.

\subsection{Reaching the Limits: Impact of Online Velocity Replanning}
This section evaluates the performance of our framework when the vehicle operates near its dynamic limits. In particular, we couple the PP and CL controllers with the MS-NN steering module, and we evaluate the effect of the online velocity replanner (FBGA, Fig.~\ref{fig:framework_overview}). All experiments are carried out on track B, and we compare the performance of different configurations with running laps.

As discussed in Sec. \ref{sec:raceline_generation}, without FBGA, the framework reaches its limits when the \textit{nominal} \GGV{} is used for raceline generation (Sec. \ref{sec:nominal_ggv}). Introducing the FBGA velocity replanner extends the feasible performance region, enabling the use of the \textit{expanded} \GGV{} envelope (Fig. \ref{fig:limit_ggv}).

\begin{table*}[h!]
\centering
\caption{Impact of online velocity replanning with FBGA on the closed-loop performance on track B. FBGA enables the use of the extended $g$-$g$-$v$ constraints, leading to improved lap times. Best results are highlighted, second best are underlined.}
\begin{tabular}{c|c|c|c|c|c|c}
\toprule
Lateral Controller & Online Velocity Replanning & $g$-$g$-$v$ Type & Max $v$ [m/s] & Max $|a_x|$ [m/s$^2$] & Max $|a_y|$ [m/s$^2$] & Lap Time [s] \\
\midrule
\multirow{4}{*}{CL + MS-NN} 
& no & nominal   & 4.001 & 2.096 & 5.915 & 6.842 \\
& no & extended  & 4.120 & 2.213 & 7.097 & -- (crash) \\
& yes (FBGA) & nominal  & 4.277 & 2.110 & 6.151 & 6.626 \\
& yes (FBGA) & extended & 4.301 & 2.585 & 6.345 & 6.449 \\ 
\midrule
\multirow{4}{*}{PP + MS-NN} 
& no & nominal   & 4.007 & 2.671 & 6.639 & 6.643 \\
& no & extended  & 3.807 & 2.815 & 7.387 & -- (crash) \\
& yes (FBGA) & nominal  & \underline{4.470} & \underline{2.838} & \underline{8.169} & \underline{6.226} \\
& yes (FBGA) & extended & \textbf{4.522} & \textbf{3.018} & \textbf{9.050} & \textbf{6.089} \\
\bottomrule
\end{tabular}
\label{tab:fbga_results}
\end{table*}

\begin{figure}[]
    \centering
    \includegraphics[width=0.46\textwidth]{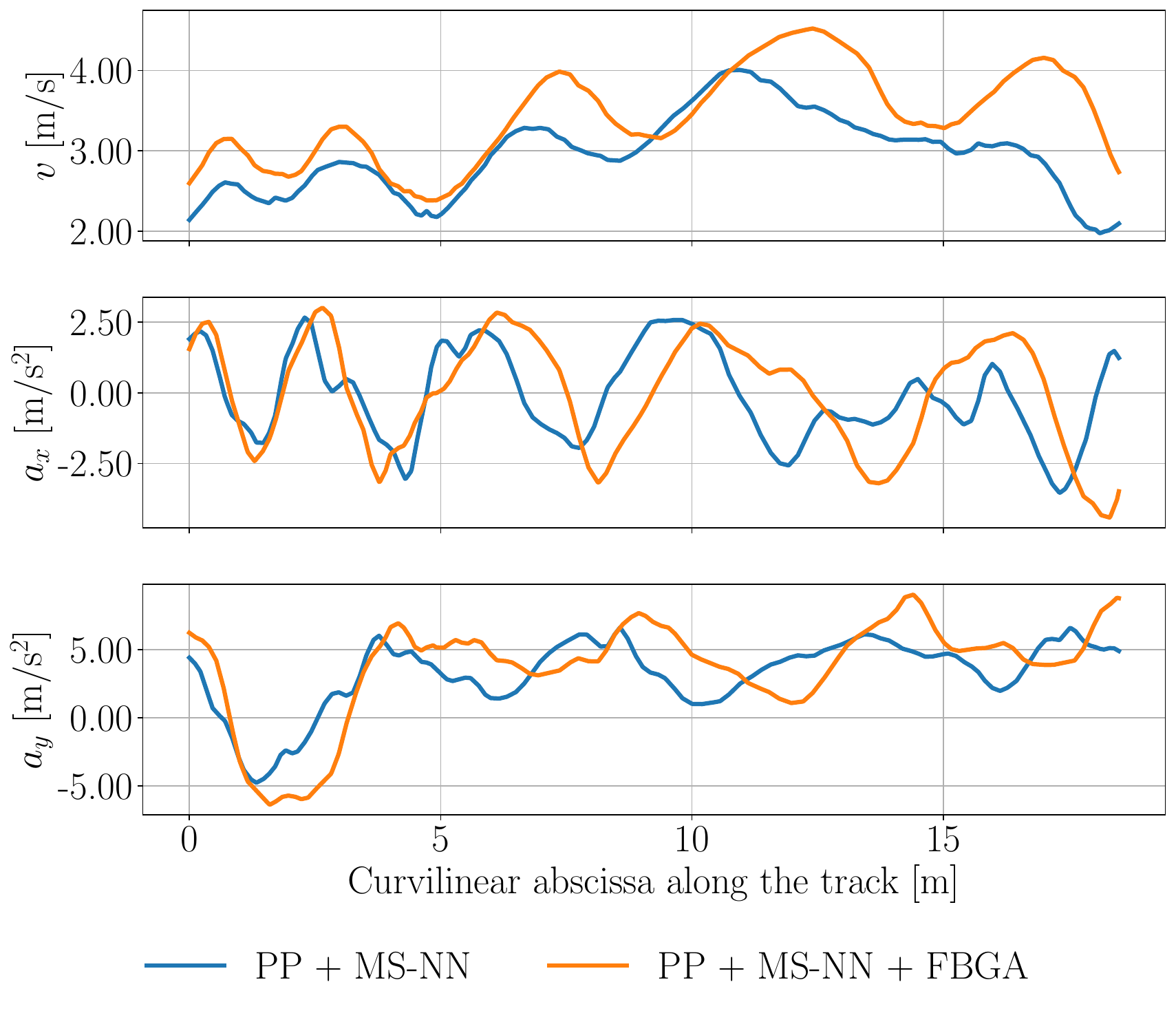}
    \caption{Velocity $v$, longitudinal and lateral accelerations $\{a_x, a_y\}$ for PP + MS-NN (nominal \GGV{}) and PP + MS-NN + FBGA (extended \GGV{}). With online velocity replanning (FBGA), the extended \GGV{} enables the vehicle to reach higher accelerations and speeds, yielding faster lap times (Table~\ref{tab:fbga_results}).}
    \label{fig:PP-MSNN_vs_PP-MSNN-FBGA}
\end{figure}

\begin{figure}[]
    \centering
    \includegraphics[width=0.90\linewidth]{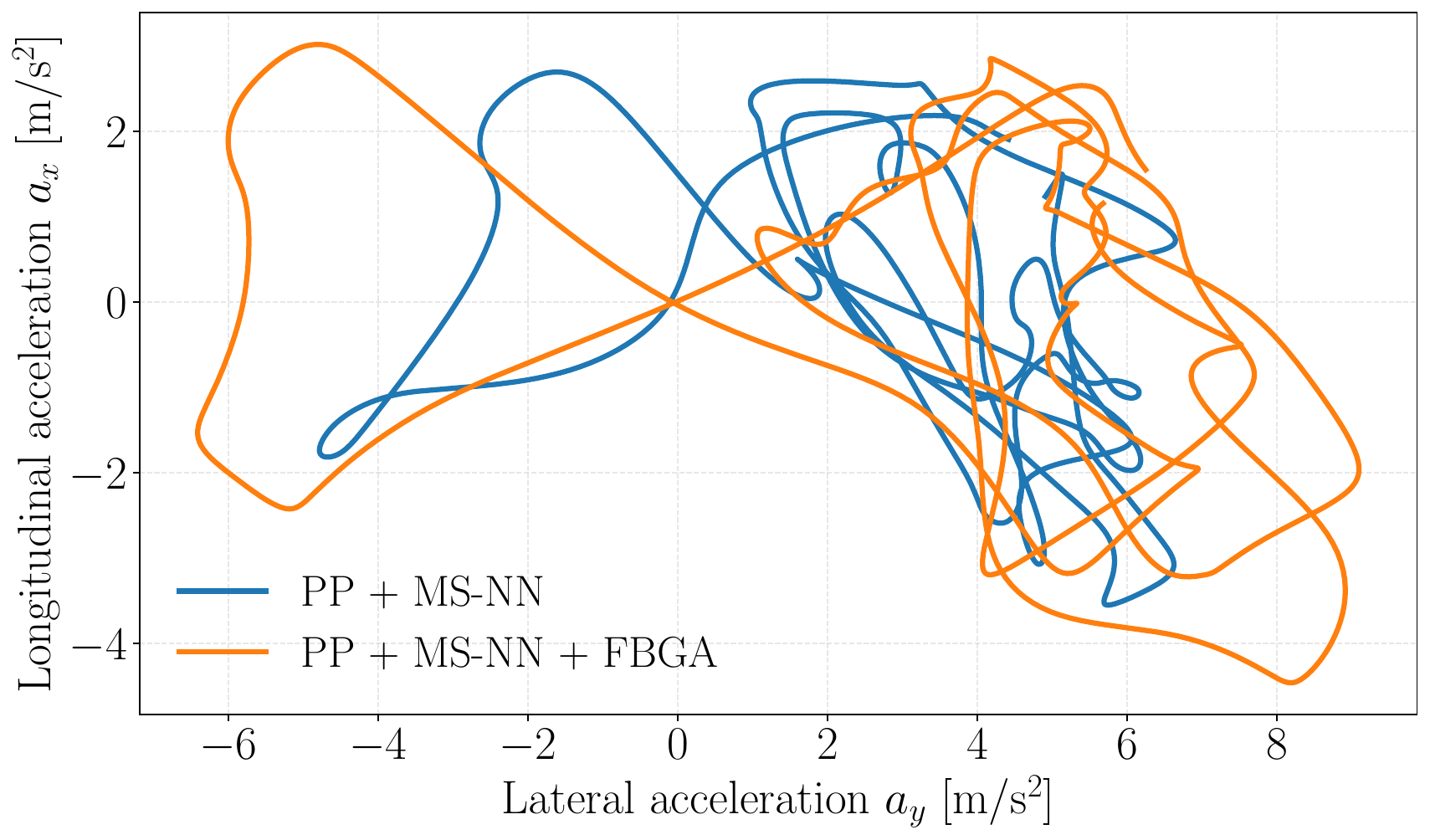}
    \caption{Executed longitudinal and lateral accelerations. PP + MS-NN follows a slower raceline under nominal \GGV{} constraints, while PP + MS-NN + FBGA reaches higher accelerations enabled by the extended \GGV{} constraints and online velocity replanning (Table~\ref{tab:fbga_results}).}
    \label{fig:Comparison_ay_ax_Lap2}
\end{figure}
\begin{figure}[]
    \centering
    \includegraphics[width=0.40\textwidth]{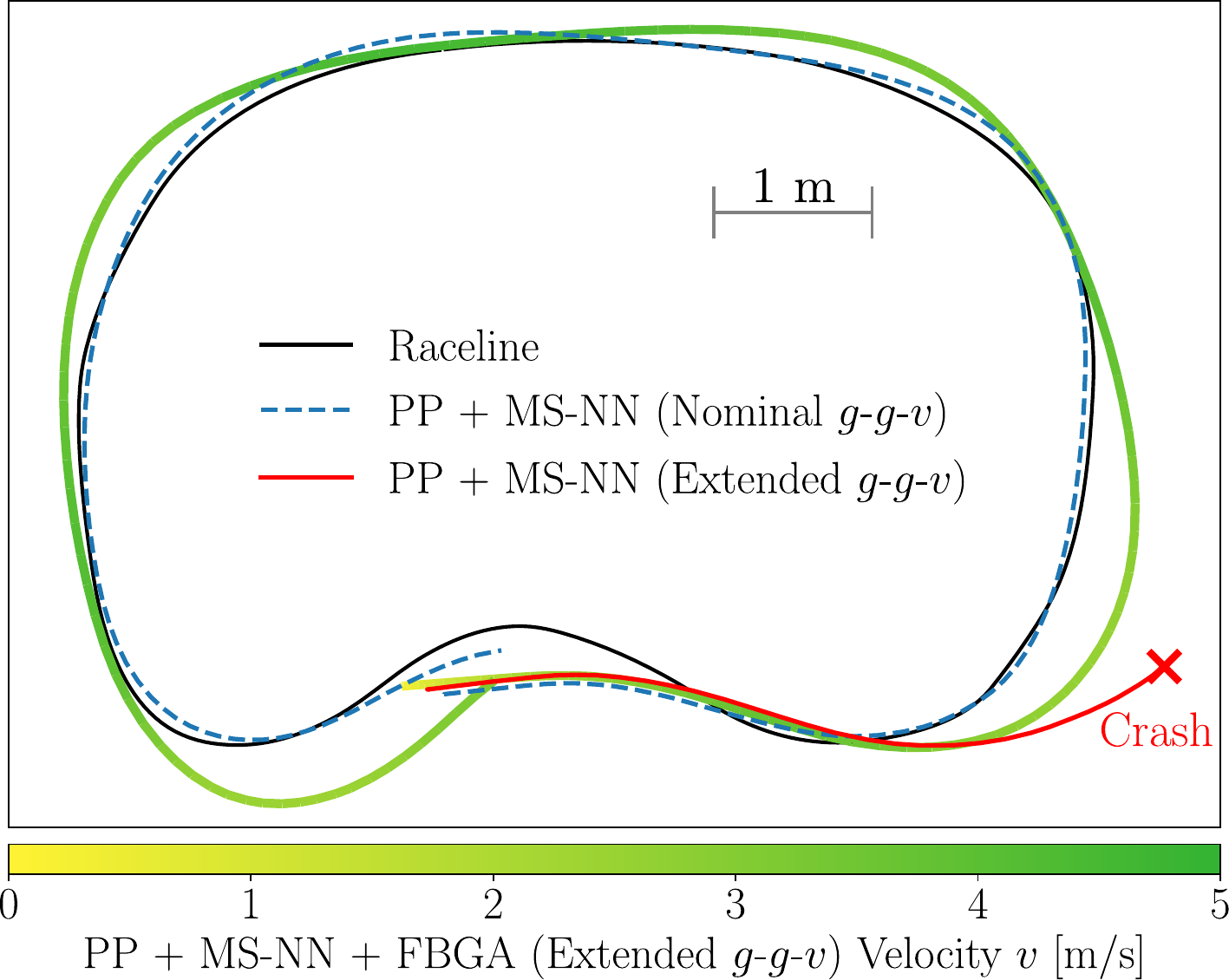}
    \caption{Trajectory of the best configuration (PP + MS-NN + FBGA) under the extended \GGV{} constraints, with the colorbar showing the velocity profile. PP + MS-NN under nominal \GGV{} (dashed) stays closer to the offline raceline but loses 8.3\% in lap time. Using the extended \GGV{} without FBGA leads to track boundary crashes (red).}
    \label{fig:Trajectory_PP_TrackB_ggv1_colorbar}
\end{figure}

As shown in Table~\ref{tab:fbga_results}, the overall \textbf{best performance} is achieved by the PP + MS-NN + FBGA configuration under the extended \GGV{} envelope. This setup improves the lap time by 0.554 s (8.3\%) compared to PP + MS-NN without FBGA. As shown in Fig.~\ref{fig:PP-MSNN_vs_PP-MSNN-FBGA} and \ref{fig:Comparison_ay_ax_Lap2}, online velocity replanning enables the vehicle to reach higher peak longitudinal and lateral accelerations (9.050 vs. 6.639 m/s$^2$) and speeds (4.522 vs. 4.007 m/s), directly contributing to the lap-time gain. Interestingly, velocity replanning also changes the driven maneuvers (Fig. \ref{fig:Trajectory_PP_TrackB_ggv1_colorbar}), with larger deviations from the offline raceline. Indeed, the replanned velocity profile adapts to local mismatches and execution errors, indirectly affecting the path. Moreover, operating with FBGA and the extended \GGV{} constraints pushes the vehicle closer to its physical limits, increasing sim-to-real discrepancies but improving lap times.

In contrast, applying the extended \GGV{} envelope without FBGA causes both PP + MS-NN and CL + MS-NN to crash into the track boundaries (Table~\ref{tab:fbga_results}, Fig.~\ref{fig:Trajectory_PP_TrackB_ggv1_colorbar}). The controllers attempt to follow the more aggressive raceline, but without velocity adaptation they cannot maintain dynamic feasibility.

Online FBGA velocity replanning also improves performance under the nominal \GGV{} constraints, with lap time reductions of 0.417 s (6.3\%) for PP + MS-NN and 0.216 s (3.2\%) for CL + MS-NN, compared to their counterparts without FBGA. Using the extended \GGV{} envelope with FBGA yields further gains: an additional 0.137 s (2.2\%) for PP + MS-NN and 0.177 s (2.7\%) for CL + MS-NN, compared to their nominal-\GGV{} configurations with FBGA.

The PP-based setups consistently outperform the CL versions. Despite the curvature-continuous nature of the CL controller, the PP configurations achieve faster lap times, indicating more effective interaction with the MS-NN and FBGA modules, and better suitability to the track geometry and vehicle dynamics in our experimental setup.

The proposed MS-NN steering controller generalizes to the aggressive racelines (with nominal and extended \GGV{}) never seen during training, delivering lower lap times and tracking errors compared to the geometric PP and CL controllers alone.
Overall, combining MS-NN with FBGA-based online velocity replanning improves closed-loop performance, allowing the vehicle to exploit a larger share of its dynamic envelope, and demonstrating suitability for autonomous racing.
\subsection{Computational Times}
\begin{figure}[]
\centering
\includegraphics[width=0.47\textwidth]{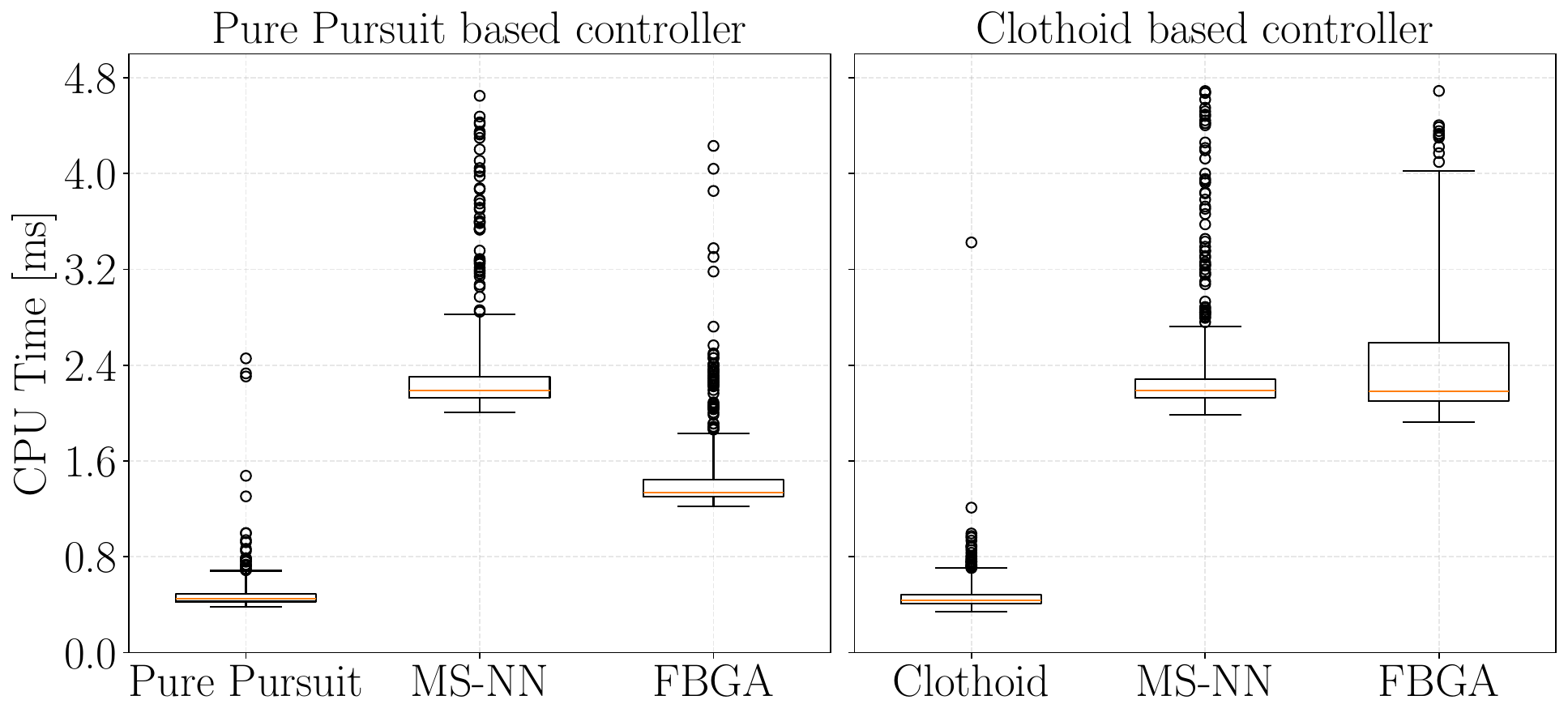}
\caption{Computational times of the main modules in our framework, running in real-time on an NVIDIA Jetson Orin Nano on the RoboRacer vehicle.}
\label{fig:cpu_times_boxplot}
\end{figure}

Our full framework runs in real time on the NVIDIA Jetson Orin Nano onboard the RoboRacer (Sec.~\ref{sec:experimental_setup}).
Fig.~\ref{fig:cpu_times_boxplot} reports the CPU times over three laps, for the PP- and CL-based configurations. The PP and CL controllers require similar computation ($\approx$0.5 ms mean). The MS-NN's inference averages 2.3 ms, with rare peaks up to 5.5 ms. FBGA is faster with PP ($\approx$1.4 ms mean), and slower with CL (2$-$4.5 ms) due to the more complex $G^2$-clothoid curvature profiles.

Considering that our whole software stack runs at 150~Hz (6.7 ms), the mean cumulative execution time is below this limit for all configurations. In rare overruns, the last control inputs are held; if the execution exceeds a 10 ms timeout, 
the system skips to the next control cycle. 
Hence, all modules meet real-time constraints on the embedded platform, with PP-based variants showing lower CPU load than CL-based ones.


\section{Conclusions}
\label{sec:conclusion}
We presented a real-world autonomous racing stack, benchmarking existing and new planning and control methods. Our framework integrates time-optimal raceline generation, online geometric path tracking (PP, CL), neural steering control (MS-NN), and optimization-based velocity replanning (FBGA). We also introduced a new MS-NN architecture for steering control under high-acceleration conditions.

We deployed the framework on a 1:10-scale RoboRacer across two tracks. Three racelines with increasing aggressiveness were generated using different \GGV{} envelopes. With a cautious \GGV{} and no FBGA velocity replanning, PP + MS-NN improved path tracking by 21.2\% and reduced steering oscillations by 36.4\% compared to PP alone. For the most aggressive raceline (extended \GGV{}), FBGA velocity replanning was required to preserve dynamic feasibility and avoid crashes. In this setting, PP + MS-NN + FBGA achieved the best results, improving lap times by 8.3\% and increasing speeds by 11.3\% over PP + MS-NN with nominal \GGV{} constraints.
Our MS-NN outperformed an existing baseline by 22.7\% FVU, showing better generalization to unseen racelines while retaining physical interpretability. Our full stack runs in real time on the embedded platform. Our open-source code and datasets provide the community with a benchmark for research in real-world autonomous racing.


\bibliographystyle{IEEEtran}
\bibliography{root} 

@article{Heilmeier2020,
author = {Alexander Heilmeier and Alexander Wischnewski and Leonhard Hermansdorfer and Johannes Betz and Markus Lienkamp and Boris Lohmann},
title = {Minimum curvature trajectory planning and control for an autonomous race car},
journal = {Vehicle System Dynamics},
volume = {58},
number = {10},
pages = {1497--1527},
year = {2020},
publisher = {Taylor \& Francis},
doi = {10.1080/00423114.2019.1631455},
eprint = {
        https://doi.org/10.1080/00423114.2019.1631455
}
}

@article{Bertolazzi2018,
title = {On the G2 Hermite Interpolation Problem with clothoids},
journal = {Journal of Computational and Applied Mathematics},
volume = {341},
pages = {99-116},
year = {2018},
issn = {0377-0427},
doi = {https://doi.org/10.1016/j.cam.2018.03.029},
author = {Enrico Bertolazzi and Marco Frego},
}

@article{Christ2021,
author = {Fabian Christ and Alexander Wischnewski and Alexander Heilmeier and Boris Lohmann},
title = {Time-optimal trajectory planning for a race car considering variable tyre-road friction coefficients},
journal = {Vehicle System Dynamics},
volume = {59},
number = {4},
pages = {588--612},
year = {2021},
publisher = {Taylor \& Francis},
doi = {10.1080/00423114.2019.1704804},
eprint = {https://doi.org/10.1080/00423114.2019.1704804
}
}

@techreport{coulter1992implementation,
  title={Implementation of the Pure Pursuit Path Tracking Algorithm},
  author={Coulter, R. Craig},
  institution={Carnegie Mellon University},
  year={1992},
  type={Technical Report},
  url={https://www.ri.cmu.edu/pub_files/pub4/coulter_r_craig_1992_1/coulter_r_craig_1992_1.pdf}
}

@article{dalio2020mental,
  author  = {Mauro Da Lio and Riccardo Don\`{a} and Gastone Pietro Rosati Papini and Francesco Biral and Henrik Svensson},
  title   = {A Mental Simulation Approach for Learning Neural-Network Predictive Control (in Self-Driving Cars)},
  journal = {IEEE Access},
  year    = {2020},
  volume  = {8},
  pages   = {192041--192064},
  doi     = {10.1109/ACCESS.2020.3032780}
}

@article{Piazza2025,
  author={Piazza, Mattia and Piccinini, Mattia and Taddei, Sebastiano and Biral, Francesco and Bertolazzi, Enrico},
  journal={IEEE Robotics and Automation Letters}, 
  title={Real-Time Velocity Profile Optimization for Time-Optimal Maneuvering With Generic Acceleration Constraints}, 
  year={2026},
  volume={11},
  number={2},
  pages={1674-1681},
  doi={10.1109/LRA.2025.3643297}
}

@article{betz2022autonomous,
  title={Autonomous vehicles on the edge: A survey on autonomous vehicle racing},
  author={Betz, Johannes and Zheng, Hongrui and Liniger, Alexander and Rosolia, Ugo and Karle, Phillip and Behl, Madhur and Krovi, Venkat and Mangharam, Rahul},
  journal={IEEE Open Journal of Intelligent Transportation Systems},
  volume={3},
  pages={458--488},
  year={2022},
  publisher={IEEE}
}

@INPROCEEDINGS{Piccinini2025ICRA,
  author={Piccinini, Mattia and Taddei, Sebastiano and Betz, Johannes and Biral, Francesco},
  booktitle={2025 IEEE International Conference on Robotics and Automation (ICRA)}, 
  title={Kineto-Dynamical Planning and Accurate Execution of Minimum-Time Maneuvers on Three-Dimensional Circuits}, 
  year={2025},
  volume={},
  number={},
  pages={1-7},
  doi={10.1109/ICRA55743.2025.11127446}}

@INPROCEEDINGS{piccinini2025model,
  author={Piccinini, Mattia and Mungiello, Aniello and Jank, Georg and Papini, Gastone Pietro Rosati and Biral, Francesco and Betz, Johannes},
  booktitle={2025 IEEE 28th International Conference on Intelligent Transportation Systems (ITSC)}, 
  title={Model-Structured Neural Networks to Control the Steering Dynamics of Autonomous Race Cars}, 
  year={2025},
  volume={},
  number={},
  pages={4129-4136},
  doi={10.1109/ITSC60802.2025.11423721}
}

@ARTICLE{subosits2021impacts,
	author={Subosits, J. Gerdes, C.},
	journal={IEEE Transactions on Intelligent Vehicles}, 
	title={Impacts of Model Fidelity on Trajectory Optimization for Autonomous Vehicles in Extreme Maneuvers}, 
	year={2021},
	volume={6},
	number={3},
	pages={546-558},
	doi={10.1109/TIV.2021.3051325}}

@Article{Bozdogan1987,
author={Bozdogan, Hamparsum},
title={Model selection and Akaike's Information Criterion (AIC): The general theory and its analytical extensions},
journal={Psychometrika},
year={1987},
month={Sep},
day={01},
volume={52},
number={3},
pages={345-370},
issn={1860-0980},
doi={10.1007/BF02294361},
}

@book{Guiggiani2018,
	author = {Guiggiani, M.},
	isbn = {9783030103354},
	publisher = {Springer},
	title = {The Science of Vehicle Dynamics: Handling, Braking, and Ride of Road and Race Cars},
	year = {2018}}

@ARTICLE{piccinini2025road,
  author={Piccinini, Mattia and Zumerle, Matteo and Betz, Johannes and Pietro Rosati Papini, Gastone},
  journal={IEEE Open Journal of Intelligent Transportation Systems}, 
  title={A Road Friction-Aware Anti-Lock Braking System Based on Model-Structured Neural Networks}, 
  year={2025},
  volume={6},
  number={},
  pages={522-536},
  doi={10.1109/OJITS.2025.3563347}
}

@article{Kabzan2019,
	author = {J. {Kabzan} and L. {Hewing} and A. {Liniger} and M. N. {Zeilinger}},
	journal = {IEEE Robotics and Automation Letters},
	number = {4},
	pages = {3363-3370},
	title = {Learning-Based Model Predictive Control for Autonomous Racing},
	volume = {4},
	year = {2019}}

@article{Wurman2022,
	author = {Wurman, Peter and Barrett, Samuel and Kawamoto, Kenta and MacGlashan, James and Subramanian, Kaushik and Walsh, Thomas and Capobianco, Roberto and Devlic, Alisa and Eckert, Franziska and Fuchs, Florian and Gilpin, Leilani and Khandelwal, Piyush and Kompella, Varun and Lin, HaoChih and MacAlpine, Patrick and Oller, Declan and Seno, Takuma and Sherstan, Craig and Thomure, Michael and Kitano, Hiroaki},
	year = {2022},
	month = {02},
	pages = {223-228},
	title = {Outracing champion Gran Turismo drivers with deep reinforcement learning},
	volume = {602},
	journal = {Nature},
	doi = {10.1038/s41586-021-04357-7}
}

@article{Kebbati2023,
author = {Kebbati, Yassine and Ait-Oufroukh, Naima and Ichalal, Dalil and Vigneron, Vincent},
title = {Lateral control for autonomous wheeled vehicles: A technical review},
journal = {Asian Journal of Control},
volume = {25},
number = {4},
pages = {2539-2563},
doi = {https://doi.org/10.1002/asjc.2980},
eprint = {https://onlinelibrary.wiley.com/doi/pdf/10.1002/asjc.2980},
year = {2023}
}

@inproceedings{Devineau2018,
	author = {Devineau, Guillaume and Polack, Philip and Altch{\'e}, Florent and Moutarde, Fabien},
	booktitle = {2018 IEEE International Conference on Intelligent Transportation Systems (ITSC)},
	doi = {10.1109/ITSC.2018.8570020},
	month = {11},
	pages = {642-649},
	title = {Coupled Longitudinal and Lateral Control of a Vehicle using Deep Learning},
	year = {2018}}

@article{AbdElmoniem2020,
author = {Ahmed AbdElmoniem and Ahmed Osama and Mohamed Abdelaziz and Shady A Maged},
title ={A path-tracking algorithm using predictive Stanley lateral controller},
journal = {International Journal of Advanced Robotic Systems},
volume = {17},
number = {6},
pages = {1729881420974852},
year = {2020},
doi = {10.1177/1729881420974852},
eprint = { 
        https://doi.org/10.1177/1729881420974852
}
}

@INPROCEEDINGS{Ghignone2025,
  author={Ghignone, Edoardo and Baumann, Nicolas and Hu, Cheng and Wang, Jonathan and Xie, Lei and Carron, Andrea and Magno, Michele},
  booktitle={2025 IEEE International Conference on Robotics and Automation (ICRA)}, 
  title={RLPP: A Residual Method for Zero-Shot Real-World Autonomous Racing on Scaled Platforms}, 
  year={2025},
  volume={},
  number={},
  pages={16664-16670},
  doi={10.1109/ICRA55743.2025.11127278}}

@INPROCEEDINGS{Vazquez2020,
  author={Vázquez, José L. and Brühlmeier, Marius and Liniger, Alexander and Rupenyan, Alisa and Lygeros, John},
  booktitle={2020 IEEE/RSJ International Conference on Intelligent Robots and Systems (IROS)}, 
  title={Optimization-Based Hierarchical Motion Planning for Autonomous Racing}, 
  year={2020},
  volume={},
  number={},
  pages={2397-2403},
  doi={10.1109/IROS45743.2020.9341731}}

@INPROCEEDINGS{Rowold2023,
	author={Rowold, Matthias and Ögretmen, Levent and Kasolowsky, Ulf and Lohmann, Boris},
	booktitle={2023 IEEE Intelligent Vehicles Symposium (IV)}, 
	title={Online Time-Optimal Trajectory Planning on Three-Dimensional Race Tracks}, 
	year={2023},
	volume={},
	number={},
	pages={1-8},
	doi={10.1109/IV55152.2023.10186701}}

@conference{Gottschalk2024,
  author={Gottschalk, Simon and Gerdts, Matthias and Piccinini, Mattia.},
  title={Reinforcement Learning and Optimal Control: A Hybrid Collision Avoidance Approach},
  booktitle={Proceedings of the 10th International Conference on Vehicle Technology and Intelligent Transport Systems - VEHITS},
  year={2024},
  pages={76-87},
  publisher={SciTePress},
  organization={INSTICC},
  doi={10.5220/0012569800003702},
  isbn={978-989-758-703-0},
  issn={2184-495X},
}

@article{Piccinini_2024_how_optimal,
  author = {Mattia Piccinini and Sebastiano Taddei and Edoardo Pagot and Enrico Bertolazzi and Francesco Biral},
  title = {How optimal is the minimum-time manoeuvre of an artificial race driver?},
  journal = {Vehicle System Dynamics},
  volume = {63},
  number = {12},
  pages = {2213--2240},
  year = {2025},
  publisher = {Taylor \& Francis},
  doi = {10.1080/00423114.2024.2407176},
}

@article{piccinini2023predictive,
  title={A predictive neural hierarchical framework for on-line time-optimal motion planning and control of black-box vehicle models},
  author={Piccinini, Mattia and Larcher, Matteo and Pagot, Edoardo and Piscini, Davide and Pasquato, Leone and Biral, Francesco},
  journal={Vehicle system dynamics},
  volume={61},
  number={1},
  pages={83--110},
  year={2023},
  publisher={Taylor \& Francis}
}

@ARTICLE{Piccinini2023_physics_driven,
  author={Piccinini, Mattia and Taddei, Sebastiano and Larcher, Matteo and Piazza, Mattia and Biral, Francesco},
  journal={IEEE Access}, 
  title={A Physics-Driven Artificial Agent for Online Time-Optimal Vehicle Motion Planning and Control}, 
  year={2023},
  volume={11},
  number={},
  pages={46344-46372},
  doi={10.1109/ACCESS.2023.3274836}}

@ARTICLE{Jahncke2026_diff_MPC,
  author={Jahncke, Felix and Zarrouki, Baha and Piccinini, Mattia and D'sa, Jovin and Isele, David and Bae, Sangjae and Betz, Johannes},
  journal={IEEE Robotics and Automation Letters}, 
  title={Differentiable Weights-Varying Nonlinear MPC via Gradient-Based Policy Learning: An Autonomous Vehicle Guidance Example}, 
  year={2026},
  volume={11},
  number={3},
  pages={3724-3731},
  doi={10.1109/LRA.2026.3662644}
}

@INPROCEEDINGS{Langmann2025,
  author={Langmann, Alexander and Ögretmen, Levent and Werner, Frederik and Betz, Johannes},
  booktitle={2025 IEEE Intelligent Vehicles Symposium (IV)}, 
  title={Online Velocity Profile Generation and Tracking for Sampling-Based Local Planning Algorithms in Autonomous Racing Environments}, 
  year={2025},
  volume={},
  number={},
  pages={632-639},
  doi={10.1109/IV64158.2025.11097653}}

@article{frego2017semi,
  title={Semi-analytical minimum time solutions with velocity constraints for trajectory following of vehicles},
  author={Frego, Marco and Bertolazzi, Enrico and Biral, Francesco and Fontanelli, Daniele and Palopoli, Luigi},
  journal={Automatica},
  volume={86},
  pages={18--28},
  year={2017},
  publisher={Elsevier}
}
	
\end{document}